\def\year{2022}\relax
\documentclass[letterpaper]{article} 
\usepackage{aaai22}  
\usepackage{times}  
\usepackage{helvet}  
\usepackage{courier}  
\usepackage[hyphens]{url}  
\usepackage{graphicx} 
\usepackage{natbib}  
\usepackage{caption} 
\DeclareCaptionStyle{ruled}{labelfont=normalfont,labelsep=colon,strut=off} 
\usepackage{algorithm}
\usepackage{algorithmic}

\usepackage{newfloat}
\usepackage{listings}
\floatstyle{ruled}
\newfloat{listing}{tb}{lst}{}
\floatname{listing}{Listing}
\title{Embedding Privacy in Computational Social Science and Artificial Intelligence Research}
\author{
    Keenan Jones\textsuperscript{\rm 1},
    Fatima Zahrah\textsuperscript{\rm 2}
    Jason R.C. Nurse\textsuperscript{\rm 1}
}
\affiliations{
    \textsuperscript{\rm 1}School of Computing \& Institute of Cyber Security for Society (iCSS), University of Kent, UK\\
    \textsuperscript{\rm 2}School of Computer Science, University of Bristol, UK\\

    j.r.c.nurse@kent.ac.uk
}

\begin{document}

\maketitle

\begin{abstract}
Privacy is a human right. It ensures that individuals are free to engage in discussions, participate in groups, and form relationships online or offline without fear of their data being inappropriately harvested, analyzed, or otherwise used to harm them. Preserving privacy has emerged as a critical factor in research, particularly in the computational social science (CSS), artificial intelligence (AI) and data science domains, given their reliance on individuals’ data for novel insights. The increasing use of advanced computational models stands to exacerbate privacy concerns because, if inappropriately used, they can quickly infringe privacy rights and lead to adverse effects for individuals---especially vulnerable groups---and society. We have already witnessed a host of privacy issues emerge with the advent of large language models (LLMs), such as ChatGPT, which further demonstrate the importance of embedding privacy from the start. This article contributes to the field by discussing the role of privacy and the issues that researchers working in CSS, AI, data science and related domains are likely to face. It then presents several key considerations for researchers to ensure participant privacy is best preserved in their research design, data collection and use, analysis, and dissemination of research results.
\end{abstract}

\section{Introduction} 

As society has advanced, the types of data available have expanded immensely. This is a consequence of  technologies such as social media, the internet of things (IoT) and artificial intelligence (AI), which have fueled an interest in gathering and analyzing this data to better understand large-scale social phenomena. The field of Computational Social Science (CSS) is pertinent to these discussions as it involves the development and application of computational techniques (such as advanced machine learning algorithms) to explore human behavioral data~\cite{lazer2020computational}. CSS allows large human-related datasets to be analyzed and can produce insights into topics including population habits and opinions, online activism, the spread of misinformation, and online hate and radicalism. Another field that has rapidly gained traction in the last three years is AI and particularly, Generative AI (GenAI), i.e., AI algorithms that can generate new content (e.g., text, audio, visuals, code). The releases of OpenAI’s ChatGPT and Dall-E 2 in 2022 were significant milestones as they demonstrated how large language models (LLMs), trained on immense datasets, could be used by anyone for a range of novel tasks, both constructive and malign. Since then, other GenAI platforms have been launched including Microsoft’s Copilot and Google’s Gemini.

While there are numerous benefits to the application of computational techniques to individuals’ data, there are also a plethora of ethical and privacy concerns that must be considered. These stem from the reality that inappropriately applying these techniques can cause immense harm to the communities they study and to the wider research domain. Ethical practices aim to ensure that research is conducted to the highest standard; ensuring that the data used has been legitimately gathered (e.g., using informed consent and obeying platform rules), accurately processed (e.g., absent of bias) and responsibly reported on (e.g., acknowledging the ethical implications of any findings). Privacy, in particular, is a core concern that emerges in any discussion involving human or social data. In CSS and AI research, this centers on maintaining the privacy of individuals whose data is used in studies and ensuring  participation does not lead to harm. 

There are multiple examples that demonstrate the relevance of privacy in such research, but some of the most topical center on the protection of personal data as it is collected, stored and analyzed, and raises questions around robustly anonymizing datasets (as shown in the well-known Netflix example~\cite{narayanan2008robust}). In AI research, the privacy challenge is similar, but also originates from the fact that the LLMs---which power ChatGPT and other systems---are often trained using significant amounts of online data including personal information. ChatGPT has already suffered a (temporary) ban in Italy over privacy fears~\cite{BBC2023ban} and Britain’s data protection regulator has recently launched a consultation into the legality of web scraping data to train generative AI models~\cite{ICO2024ai}. 

This article contributes to the discussion by providing a foundation for CSS and AI researchers in conducting privacy-aware studies. While our focus is on CSS, we also consider related advancements in AI and data science considering their prominence and reliance on individuals’ data. To achieve this, we first reflect on how privacy has been discussed to date. We further engage in our own critical discussion about the privacy challenges that arise in working with data, and ensuring that aggregated data and any subsequent analyses/releases do not lead to adverse effects. To provide guidance for researchers, we then conclude by presenting a number of recommendations addressing how privacy should be planned for and embedded in CSS and AI research.

\section{The Role of Privacy in Research} 

\subsection{Privacy and Information Control}

At its core, privacy has typically been regarded as an individual’s ‘right to be left alone’~\cite{Decew2018priv}, and in light of this, attempts have been made to examine how privacy should be applied to the evolving nature of data in the digital age. Privacy is also framed as empowerment, and therefore that individuals should have the power to control and monitor the distribution of their personal data online~\cite{langheinrich2005living}. While offering a simple extension to traditional conceptions of privacy~\cite{Decew2018priv}, this approach brings with it further complications. Once data is released onto the web it is all but impossible to ensure its removal, and this raises questions regarding the extent to which an individual can truly maintain control over their information online. If a person has allowed a given web platform, for instance, to publish their posts, can they ever truly regain control over this information? Currently, it is difficult to provide a concrete answer to this question. 

What is clear, however, is that in the digital age a vast degree of responsibility towards the handling of personal data, including the role of privacy protection, has been ceded to those controlling the data~\cite{barth2017privacy}. This is typically viewed in terms of the large companies who hold the monopoly over this form of data, but within this bracket also fall the CSS and AI researchers. In conducting their studies,  researchers place themselves in a position of control over vast swathes of personal data, and by extension the privacy of the individuals included in these datasets. 

\subsection{The Direct Dangers of Poor Privacy Consideration}

Beyond the broader conceptions of privacy as integral to preserving one’s inherent rights, it is also essential to consider the ways in which poor considerations of privacy can lead to tangible harm. There are at least two key threats to user privacy directly relevant to research that warrant discussion~\cite{katal2013big}. 

Firstly, there is the threat of direct inference, in which a malicious actor combines publicly available personal information about a given individual (extracted, say, from their social media account) with information captured in a given dataset. Through this combination, the malicious actor may then be able to infer further characteristics or information about the individual that they had intended to keep secret. The second threat is that of indirect inference via the predictive power contained in large amounts of data. Through the curation of these vast datasets, it has now become commonplace to use powerful machine learning algorithms to predict personal attributes and behaviors. By leveraging these models, it is possible to infer private aspects of an individual from more “innocent” data such as social media posts (e.g.,~\cite{nouh2019understanding}. 

Research that leverages these approaches can potentially intrude on user privacy in harmful ways, either directly through the creation of these models as part of their research, or indirectly in the case of malicious actors utilizing the data collected during research to build their own privacy-busting predictive models~\cite{katal2013big}. In recent years we have seen these dangers realized in a variety of cases, including the infamous Facebook-Cambridge Analytica scandal in which Facebook data used to psychologically profile users was leveraged to manipulate opinions on the 2016 US presidential elections and the UK Brexit referendum~\cite{isaak2018user}. 

\subsection{The Problem of Privacy and Mass Data Gathering}

There are several issues relevant to the alignment of participant privacy with the research design. Arguably most central to this alignment is the principle of “informed consent”, an issue which has formed the backbone of most discussions of ethics and privacy in research involving human participants. The importance of informed consent is emphasized by most scholarly societies, including the British Society of Sociologists (BSA), the British Psychology Society (BPS), and the British Society of Criminology (BSC), who  in their framework of ethics highlight that individuals should be able to take part in research “voluntarily, free from any concern…able to give freely informed consent in all but exceptional circumstances”~\cite{BSC2015ethics}. 

Despite the initial clarity of the role of informed consent in the analysis of personal data, it is arguably immediately compromised when applied to the large datasets common in CSS, which are typically collected at scale, and often from online platforms like X (Twitter), TikTok, WhatsApp, Reddit, and Facebook. As noted in the BSA’s ethical guidelines for digital research, to work with this form of data requires dealing with “new, messy and often confusing definitions of the private and the public” which require the resolution of “unprecedented tensions between the researcher and the researched”~\cite{BSA2017eth}. These tensions, combined with the massive scale typical of online data gathering, often mean that traditional approaches to gaining informed consent are fundamentally “impractical”~\cite{AoIR2019ethic}. 

To help mitigate this, in their terms of service (ToS) some platforms include descriptions of how user data may be made available for research purposes. Furthermore, attempts have been made to explore the existence of informed consent in the analysis of online data; and whether the acceptance of these ToS amount to tacit consent~\cite{fiesler2020no}. However, many studies have demonstrated that users seldom read or understand the agreements that they are making in these ToS, particularly with respect to a given platform's usage of their data~\cite{obar2020biggest}. It is, therefore, hard to claim that this can truly be understood as informed consent. 

This inability to provide informed consent necessitates a loss of control over one’s personal information. As the individual may have little understanding of how their data may be used in a given study, control over their data is largely handed to the platforms through which the data is distributed, and to those who seek to use this data. These platforms---and the third parties (of which CSS and AI researchers are an example) they share this data with---are thus able, to a great degree, to decide how long they retain this data, who this data may be shared with, and to what use this data can be put to~\cite{denardis2015internet}.

The issues covered thus far lead to a somewhat frustrating conclusion. In general, current notions of the right to privacy and of how a study can be best constructed to preserve privacy may be incompatible with the large-scale, detailed datasets central to research. Ultimately, there are many unanswered questions regarding the alignment of traditional standards of privacy, particularly in terms of consent and information control, with current trends towards digital data gathering and mass data analysis within the field. Next, we look at guidance for how this may be addressed. 

\section{Conducting Privacy-Aware Research}

Given the need to align privacy with CSS and AI research, it is essential that researchers design their projects with a mind towards participant privacy. Here we focus on CSS and AI research (with some consideration of research that \textit{uses} AI in social science as well), and examine the critical stages of a typical project that may require privacy-protecting measures: the initial research design, the collection and use of data (e.g., its storage and dissemination), and the analysis and model development process. We note that while this guidance is primarily to researchers conducting the research, it is also relevant to journal editors and conference program committees given their role as peer reviewers and gatekeepers for robust, ethical academic publications. 

\subsection{Research Design}

The first part of conducting privacy-aware CSS and AI research involves understanding the social context of the proposed research and the impact it could have on society and any participants involved. The most comprehensive way to do this is by making use of frameworks and guidelines provided by ethical review boards and legal regulatory bodies. For instance, the EU’s General Data Protection Regulation (GDPR) requires a Data Privacy Impact Assessment (DPIA) to be carried out before the start of any project, so as to ensure a privacy-by-design approach. This framework has been utilized by researchers directly~\cite{martin2018developing} and in other work researchers have focused on protection goals around data availability, unlinkability and transparency for participants’ personal data~\cite{bieker2016process}. While GDPR is cited as an example above, as readers consider this section they should note that each jurisdiction will have its own regulatory and advisory structures relevant to researchers; e.g., the US's CPRA and HIPPA, Singpore's PDPA, Japan's APPI, South Korea's PIPA and Brazil's LGPD. Therefore, \textbf{addressing the privacy concerns associated with handling digital data will require that guidance from regulatory bodies and context-specific areas (e.g., protocols for vulnerable groups such as ethnic minorities or children) are consulted}.

It is also not uncommon for researchers to look to the ToS of the various organizations and online platforms responsible for providing the data to ascertain whether and how they should interact with it~\cite{BSC2015ethics}. However, while these considerations are important and can offer some guidance in this area, they are not inherently sufficient for respecting participant privacy. As discussed prior~\cite{fiesler2020no}, there is not a guaranteed relationship between what is permitted in a platform’s ToS and what is best for an individual’s privacy. 

In sum, we can reason that although a given platform’s ToS may allow for something to be done with a user's data, this does not guarantee that that individual is meaningfully consenting to their data being used in such a way. \textbf{Therefore, although it is important to comply with all relevant ToS, this alone may be insufficient to guarantee ethical privacy preservation. Beyond this, researchers must assess the extent to which participant privacy can be maintained within their studies and consider which methods or practices would be most suited for this.} To focus specifically on AI, there is a growing body of \textbf{research into ethical and responsible AI that should be consulted when planning studies to consider topics including transparency, justice and fairness and responsibility}. 

\subsection{Data Collection and Usage}

Informed consent, in particular, has emerged as a notable problem in research involving big data, and is a major area of consideration in the initial collection of participant data. Such research uses a range of data collection techniques, including automated scraping APIs (as adopted to train many of today’s LLMs), as well as datasets collected by other researchers or organizations. As personally identifiable information or sensitive information is often being collected, strong steps are required to protect the identity of individual participants and, where possible, to obtain their informed consent to the research being carried out upon them or their data~\cite{AoIR2019ethic}. Gaining this consent is usually impracticable in the case of big data research however, and therefore researchers have been forced to rely upon different processes to ensure participant privacy is appropriately safeguarded. Some researchers try to obtain first-degree informed consent or retroactive consent from relevant participants before publishing any findings, whilst others focus on deleting names and other highly identifiable information from the dataset~\cite{iphofen2020handbook}. Ultimately, \textbf{it is important to not only use guidance from up-to-date ethical guidelines and regulation regarding participant consent, but to also reflect on examples from context-specific research to understand the privacy implications of the research for certain communities as well as for individual participants.}

As mentioned above, in addition to generating new empirical data for analysis, researchers also make use of previously collected data from other researchers or organizations. A small amount of such research has also made use of data that has been accessed and released through illicit means, including as a result of an unintended disclosure by the data owner, an unauthorized leak by someone with access to the data, or an exploitation of a vulnerability in a computer system~\cite{thomas2017ethical}. For instance, studies have been conducted using datasets released by WikiLeaks (a non-profit organization that publishes information leaks provided by anonymous sources). These datasets may be insightful but raise profound privacy concerns since the informed consent and privacy of the participants involved has already been breached. In such cases, researchers must be able to prove their intended research is of high social value, and that benefits clearly outweigh any relevant harms~\cite{ienca2021ethical}. \textbf{It is, therefore, vital to involve independent review committees to help assess risks and benefits of the research before data collection or access begins.}

Moreover, when integrating privacy protection into their data handling, researchers must also consider strategies for data storage. Here, the researcher must consider how they intend to store their data, and the safeguards they intend to put in place to ensure its protection. \textbf{At a minimum, it is crucial that all data be stored behind secured machines to minimize their access to only those intended, and, where possible, further encryption of the datasets should be used to provide added protections}~\cite{AoIR2019ethic}. In the context of GenAI systems, one additional point to note is that they are vulnerable to attacks which can expose personal data including chat messages and device information; as demonstrated in~\cite{wired2023chat}. \textbf{For GenAI systems, researchers must take care in managing access as well as third-party engagement with interactive systems---this may also mean that it may be useful to test systems for vulnerabilities before their release}.

Beyond using the collected data for their personal analyses, it is typical that researchers should seek to disseminate their datasets to the broader academic community. However, with the sharing of data comes the potential for added risk to the privacy of participants within the dataset. In order to mitigate this, the first consideration that researchers should make is the methods by which they shall anonymize their datasets prior to sharing. Where possible, \textbf{it is good practice to ensure that any recipients of the dataset are unable to identify the participants whose data they are accessing. To do this, researchers will typically make efforts to remove identifying features and pseudonymize any names or usernames present in the dataset.} These are important steps to ensuring that participant privacy is safeguarded. 

While previous research has found that many anonymization techniques can be bypassed~\cite{narayanan2008robust}, this does not inherently invalidate the use of anonymization; it can still offer some protection. This does mean nonetheless that anonymization alone may not be sufficient to safeguard privacy. Therefore, researchers should make further considerations of the risks that these adversarial de-anonymization attacks pose to participants in the dataset, and weigh up whether these risks can be justified when sharing the data. To further protect against this, \textbf{researchers should also take additional steps to vet potential data recipients, rather than releasing the data en masse to the public}, which can help further protect participants. Finally, researchers should ensure that they provide good documentation in any publications accompanying the dataset, providing clear descriptions of the steps that have been taken to protect their datasets and the participants within them. The common adage broadly applies: \textbf{“research data should be as open as possible but as closed as necessary” in order to ensure a balance is struck between availability and privacy protection}~\cite{landi2020fair}.

Researchers should also ensure that processes are established for scenarios in which their datasets are inadvertently leaked. The exact approach taken by researchers is dependent on the manner of the data being stored and the degree to which prior consent has been obtained, so careful consideration is essential to ensuring that every possible step is taken to preserve participant privacy as much as possible when storing personal data. Therefore, \textbf{it is vital that researchers consider any containment protocols that are to be used, the manner in which backups of datasets are taken and stored, and whether (and how) they intend to notify participants should their data be leaked}~\cite{AoIR2019ethic}.

\subsection{Analysis and Dissemination of Results}

Beyond data handling, it is also crucial that researchers ensure that any analysis or model creation conducted is done with due respect to participant privacy. In terms of analysis of participant data directly, it is thus essential that researchers give considerable thought to the issues posed by identification~\cite{AoIR2019ethic}. This concerns the potential for participant identities included in any datasets to be discovered based on the results of the analyses conducted. This also covers the ability of any readers to identify participants through the subsequent publication of these results. Consideration of this is crucial, as failures here could undo any attempts at anonymization made at the data gathering stage. As an example, if quotations of a given online post are published alongside a prediction of that post’s geographic location, these details could quite easily be leveraged to identify the user behind the post, even if the post itself is sanitized of identifying characteristics. \textbf{As such, it is necessary that researchers consider the extent to which individuals can be reidentified through their analyses and results publication. As far as possible, researchers should ensure that these forms of participant identification through a study’s results are not possible.}

Besides issues of identification, another consideration that researchers must make is the potential downstream impacts of their work on people’s privacy. This is particularly relevant when developing powerful models (or language models), which are trained on large amounts of user data and then made available for use in a range of tasks~\cite{bender2021dangers}. Ostensibly, it is tempting to assume that if a model is trained and released with the source training data being kept private, participant privacy is therefore maintained. However, studies have now shown that these models are prone to leaks whereby extracts of private information within the model’s training data can be reproduced or predicted with relative ease~\cite{bender2021dangers}. Moreover, the nature of many of the models developed in CSS and AI research mean that one must account for their potential privacy-busting applications. Models such as those that can predict political leaning, personality, and location, could all be leveraged to maliciously attack the privacy of other individuals not considered in the original study~\cite{brundage2018ai}. \textbf{Researchers must therefore consider and evaluate their models’ susceptibility to these downstream risks, and evaluate how their models can be safely shared in light of this.} The reality is that while models may be developed for good, they can often also be used maliciously. 

\section{Privacy Protection: A Researcher’s Responsibility}

Aligning privacy with the aims of CSS and AI research is a worthy challenge. Privacy considerations are vital at every step of a research project, and failure at any stage could result in genuine harm being exacted on participants, who are often unaware that their data is being used. As researchers, it is therefore of utmost importance to acknowledge the position of responsibility that we hold, and to ensure that any studies conducted using large-scale human or behavioral data is approached in a way that respects each individual’s reasonable expectations of privacy and shields them from harm. It is essential: that we as a community recognize the importance of these responsibilities and strive to uphold the highest standards of privacy protection in our research; that we constantly work to improve our privacy-protection strategies; that we develop all research projects with a mind towards privacy; and that each step of our research is conducted using the highest standards of privacy protection. This involves acknowledging not only the considerations highlighted in this paper, but the further considerations that present themselves across the innumerable ways in which research can be conducted.

\bibliography{main}

\end{document}